\documentclass[letterpaper, 10 pt, conference]{ieeeconf}  
\let\labelindent\relax         

\usepackage{times}
\usepackage{graphicx}
\usepackage{epstopdf}

\usepackage{mdwmath}
\usepackage{mdwtab}
\usepackage{rotating}
\usepackage{caption}
\usepackage{subcaption}

\usepackage{amssymb}
\usepackage{amsfonts}
\usepackage{amsmath}
\usepackage{bm}

\usepackage{cite}

\usepackage{multirow} 
\usepackage{multicol}
\usepackage{array}

\usepackage{standalone}
\usepackage{booktabs} 

\usepackage{siunitx} 

\usepackage{accents}

\usepackage{pgfplotstable} 
\usepackage{verbatim}
\usepackage{isomath}
\usepackage{overpic}
\usepackage{tikz}
\usetikzlibrary{shapes,calc,patterns,
	decorations.pathmorphing,
	decorations.markings}

\tikzstyle{spring}=[very thick,decorate,decoration={zigzag,pre length=2,post
	length=2,segment length=6}]

\tikzstyle{damper}=[thick,decoration={markings, 
	mark connection node=dmp,
	mark=at position 0.5 with 
	{
		\node (dmp) [thick,inner sep=0pt,transform shape,rotate=-90,minimum
		width=15pt,minimum height=3pt,draw=none] {};
		\draw [thick] ( $(dmp.north east)+(2pt,0)$ ) -- (dmp.south east) -- (dmp.south
		west) -- ( $(dmp.north west)+(2pt,0)$ );
		\draw [thick] ( $(dmp.north)+(0,-5pt)$ ) -- ( $(dmp.north)+(0,5pt)$ );
	}
}, decorate]

\makeatletter\newcommand{\manuallabel}[2]{\def\@currentlabel{#2}\label{#1}}\makeatother

\usepackage{color}
\usepackage{xcolor}

\colorlet   {lightorange}{orange!20}
\colorlet   {lightgrey}  {gray!20}

\usepackage{amssymb}
\usepackage{mathtools}

\mathchardef\mhyphen="2D   

\newcommand{\RNum}[1]{\uppercase\expandafter{\romannumeral #1\relax}}

\graphicspath{
{figures/}
}

\usepackage[inline]{enumitem}
\setlist{nolistsep}

\usepackage[normalem]{ulem}                                        
\usepackage{marginnote}
\usepackage[textwidth=10ex,colorinlistoftodos]{todonotes}

\colorlet{fwu}{red}
\colorlet{ywu}{blue}
\colorlet{xzhao}{green}
\colorlet{ywang}{purple}

\IEEEoverridecommandlockouts
\usepackage[ruled,vlined]{algorithm2e}

\usepackage{amsmath}
\usepackage[utf8]{inputenc}
\usepackage{float}
\usepackage{cuted}
\usepackage{caption}
\usepackage{threeparttablex}
\usepackage{tabularx} 
\usepackage{array}
\usepackage[inkscapelatex=false]{svg}
\usepackage{hyperref}
\usepackage{authblk}

\title{\LARGE \bf%
Video-to-BT: Generating Reactive Behavior Trees from Human Demonstration Videos for Robotic Assembly
}

\author{
Xiwei Zhao$^{1,2*}$,
Yiwei Wang$^{1*}$,
Yansong Wu$^{1}$, 
Fan Wu$^{1,3\dagger}$, 
Teng Sun$^{3}$, \\
Zhonghua Miao$^{3}$,
Sami Haddadin$^{4}$,
Alois Knoll$^{1}$
\thanks{\textsuperscript{*} Equal contribution.}
\thanks{$^\dagger$ Corresponding author: Fan Wu ({\tt\small wufan@shu.edu.cn}). }
\thanks{$^{1}$ Munich Institute of Robotics and Machine Intelligence (MIRMI), Technical University of Munich, Germany. 
$^{2}$ Aalto University, Espoo, Finland.
$^{3}$ Shanghai University, Shanghai, China.
$^{4}$ Mohamed Bin Zayed University of Artificial Intelligence, Abu Dhabi, UAE.
}
}

\renewcommand{\baselinestretch}{0.935} 

\let\oldtwocolumn\twocolumn
\renewcommand\twocolumn[1][]{%
    \oldtwocolumn[{#1}{
    \begin{center}
           \includegraphics[width=1.0\textwidth]{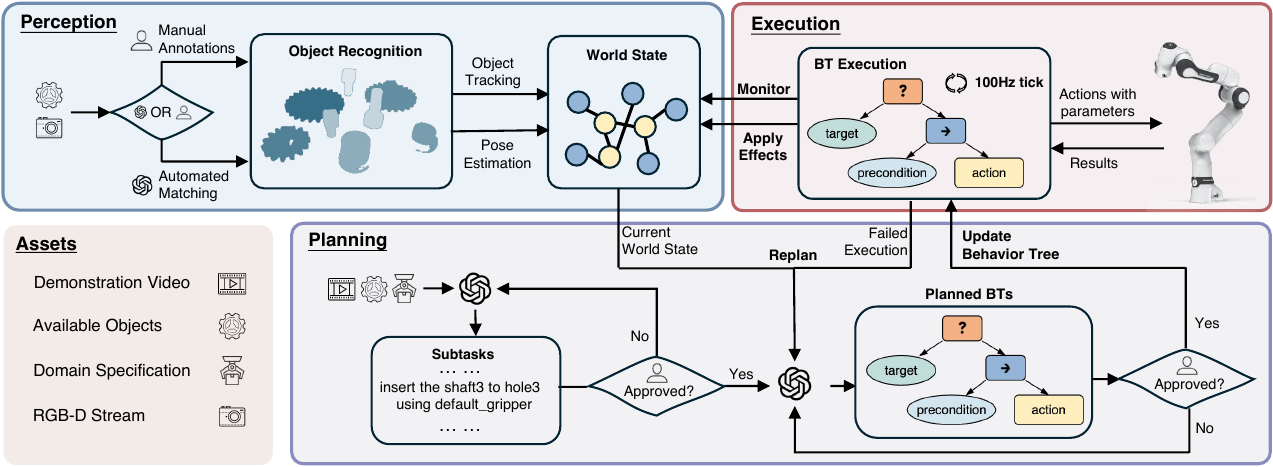}
           \captionof{figure}{\textbf{Overall framework}. The proposed framework operates through the synergy of three core modules: Planning, Perception, and Execution. Initially, the Planning Module generates BTs from video demonstrations, while the Perception Module maintains a real-time semantic understanding of the environment via the construction and update of the world state. The Execution Module then leverages the BTs and world state to command the robot's actions. This architecture forms a crucial closed loop: when the Execution Module detects a need for replanning based on new perceptual data, it triggers the Planning Module to refine the BT, ensuring robust and reactive task completion.}   \label{fig:overview}
    \end{center}
    \vspace{-0pt}
    }]
}

\begin{document}
\maketitle

\begin{abstract}
Modern manufacturing demands robotic assembly systems with enhanced flexibility and reliability. However, traditional approaches often rely on programming tailored to each product by experts for fixed settings, which are inherently inflexible to product changes and lack the robustness to handle variations.
As Behavior Trees (BTs) are increasingly used in robotics for their modularity and reactivity, we propose a novel hierarchical framework, Video-to-BT, that seamlessly integrates high-level cognitive planning with low-level reactive control, with BTs serving both as the structured output of planning and as the governing structure for execution.
Our approach leverages a Vision-Language Model (VLM) to decompose human demonstration videos into subtasks, from which Behavior Trees are generated. 
During the execution, the planned BTs combined with real-time scene interpretation enable the system to operate reactively in the dynamic environment, while VLM-driven replanning is triggered upon execution failure. This closed-loop architecture ensures stability and adaptivity. 
We validate our framework on real-world assembly tasks through a series of experiments, demonstrating high planning reliability, robust performance in long-horizon assembly tasks, and strong generalization across diverse and perturbed conditions. Project website: \href{https://video2bt.github.io/video2bt\_page/}{https://video2bt.github.io/video2bt\_page/}
\end{abstract}


\section{Introduction}

The pivotal shift from mass production to mass customization within modern, dynamic factory environments necessitates production lines that are not only reliable but also flexible~\cite{pantano2022capability,xu2025embodied,nikiforidis2025enhancing}. Specifically, the principle of inherent safety demands high execution transparency and predictability, especially as human-robot collaboration becomes commonplace. Moreover, the goal of customization demands flexibility to manage high product diversity, while the operational environment requires robustness against shop-floor disturbances like component displacements or human interventions. 
Conventional robotic systems, fundamentally reliant on static and pre-programmed tasks, are thus ill-equipped for the procedural variations and dynamic uncertainties. 
Therefore, realizing the promise of smart manufacturing calls for a new generation of robotic systems, endowed with both the cognitive intelligence for high-level task planning and reasoning, as well as the adaptive mechanisms for robust and transparent low-level execution in dynamic environments.

To design such robotic systems, they can be modeled as discrete-event dynamic systems (DEDS)~\cite{cassandras2007introduction} by abstracting the primitive action execution and state transitions as discrete events. Traditional DEDS supervisory controllers, such as Petri nets~\cite{zurawski1994petri, costelha2007modelling} and finite state machines~\cite{foukarakis2014combining, conner2017flexible}, often face scalability limitations and provide limited support for recovery from disturbances~\cite{colledanchise2018behavior}. Against this backdrop, the Behavior Trees (BTs), which represent policies in a hierarchical tree structure, have drawn increasing attention. Their modularity, human-readability, and reactivity make them well-suited to integrated planning and control in long-horizon tasks under dynamic environments. 
While there are existing methods for automated BT generation, the practical use of BTs in long-horizon tasks still relies on deterministic settings and substantial user inputs. The BT generation from human-centric instructions and subsequent application of BTs to adaptive control remain largely underexplored.

The challenge of automatically constructing BTs from human-centric instructions finds a promising solution in the recent advancements of foundation models, such as Vision-Language Models (VLMs) and Large Language Models (LLMs). 
The advanced faculties of the models for syntactic synthesis and logical inference~\cite{nascimento2024llm4ds} enable the direct generation of these abstract plans into formally specified, compositional structures such as BTs. 
Furthermore, these models possess unprecedented capabilities in common-sense reasoning and generalization, allowing them to interpret diverse, high-level instructions and decompose them into logical task sequences~\cite{huang2022inner,black2023zero}. 
While early works have explored using LLMs to generate BTs from simplified instructions~\cite{ao2025llm}, the full potential of leveraging VLMs to bridge rich, multi-modal perception directly with the generation of complex, structured BTs remains largely untapped.

To this end, we propose \textit{Video-to-BT}, a framework that combines the generalization and reasoning capabilities of foundation models with the advantages of BTs to address long-horizon assembly tasks in dynamic environments. In our approach, a VLM serves as the brain for high-level planning: it infers the task sequence from a video demonstration and generates BTs as a structured representation of the sequential manipulation plan. These BTs are then employed as a modular and reactive control architecture, executing under perception-based scene interpretation that monitors the environment states. In this way, the robot receives instructions to initiate or suspend the primitive actions. To guarantee safe execution, the framework adopts a human-in-the-loop (HITL) workflow via a web-based interface, ensuring accessibility for non-experts. The presented framework is evaluated by a real  world long-horizon assembly task.
The results demonstrate the superiority of our framework, showcasing high planning accuracy, robust execution of assembly tasks, and strong reactivity to external disturbances in dynamic environments.
To summarize, the main contributions of this work are as follows:

\begin{enumerate}

    \item We propose a novel planning framework that translates human demonstration videos into structured, executable BTs for complex robotic assembly. The key to the planning framework is a three-stage pipeline that uniquely integrates VLM-driven task decomposition, parameterized plan formulation, and automated BT synthesis.


    \item A supervisory control framework is developed that integrates a BT executor with real-time semantic perception to deliver precise, context-aware scheduling under dynamic conditions. By continuously monitoring the workspace and incorporating a recovery mechanism, the system possesses the capability to react to external disturbances.


    \item The framework is validated in a systematic way. The results demonstrate that the framework achieves accurate planning across diverse demonstrations and sustains a high completion rate for long-horizon tasks, even under disturbances, highlighting the practical reliability and robustness of the proposed methods.

\end{enumerate}

\section{Related work}

\subsection{Adaptive Robotic Assembly in Dynamic Environments}


The pursuit of enhanced flexibility for diverse tasks and robustness against environmental disturbances is a central challenge in modern smart manufacturing. While prominent approaches enhance adaptability in dynamic environments, such as Digital Twins (DTs) through high-fidelity simulation~\cite{zhang2024digital,shi2025digital} and Vision-Language-Action (VLA) models through powerful generalization~\cite{team2024octo,kim2024openvla,black2410pi0}, they both demand substantial upfront investment in terms of expert labor and resources. DTs demand intensive, expert-driven labor for creating simulators and digital assets~\cite{reif2025expert,arin2023systematic}, whereas VLAs face a critical bottleneck in their dependency on large-scale datasets~\cite{kawaharazuka2025vision}. These challenges are especially acute for complex assembly, which demands contact-rich tactile data that is notoriously scarce and difficult to acquire. To circumvent the need for extensive task-specific training or data collection, a hierarchical framework is proposed that leverages open-vocabulary foundation models to dynamically sequence a library of robust skill primitives~\cite{johannsmeier2025process}.


\subsection{VLM-based Task Planning in Robotics}
VLMs have shown significant promise for robotic task planning, due to their powerful environmental perception, common sense knowledge, and strong generalization, which enable the generation of multi-step online task plans, as seen in embodied reasoning~\cite{Ji_2025_CVPR,yang2025embodiedbench,zhang2025embodied,fang2025robix}. However, the generalization capability derived from large-scale, general-purpose datasets does not readily transfer to precision-critical tasks such as assembly. Instead of attempting to generate plans directly, a common strategy is to leverage the multi-modal understanding of VLMs to interpret human-provided instructions, such as diagrams or manuals, to derive high-level task sequences~\cite{tie2025manual2skill,li2025vlm}. Although this approach is effective, its primary limitation is the dependency on formal documentation, which precludes its use for novel or undocumented tasks. To bridge this gap, this work introduces a novel planning module that leverages a VLM to interpret human assembly demonstration videos as a rich, multi-modal source of instruction. From these demonstrations, our approach extracts nuanced procedural knowledge, offering a more robust and generalizable solution that enables learning from non-expert demonstrators and bypasses the need for formal documentation.



\subsection{Behavior Tree in Robotics}


Compared to other control architectures such as Petri nets \cite{zurawski1994petri, costelha2007modelling} and finite state machines\cite{foukarakis2014combining, conner2017flexible}, BTs offer a comparable hierarchical organization and modularity, while providing the additional advantage of reactivity\cite{colledanchise2018behavior}. Consequently, BTs have been increasingly applied in task planning and task execution.
On the planning side, existing studies have explored both analytical formulations \cite{colledanchise2019towards,scheide2025synthesizing} and learning-based approaches \cite{styrud2022combining,deng2023learning} for BT generation, yet these methods still rely heavily on pre-defined planning requirements and extensive manual programming. While there are some studies that address it by leveraging foundation models to automatically generate BTs~\cite{ao2025llm, merino2025behavior, meng2025behavior}, they still require substantial user input to specify planning requirements. 
On the control side, some works \cite{akkaladevi2024towards}\cite{wake2025vlm} have applied BTs to accomplish long-horizon tasks under fully deterministic settings, lacking adaptivity against uncertainty. In contrast, some studies \cite{Wu2024icra, cloete2024adaptive} have emphasized the adaptive execution, but reliance on handcrafted design restricts their applicability. Inspired by these works, our framework integrates the automated BT generation based on task decomposition, supervisory control for long-horizon execution, and reactive control for disturbance handling to enable robust task execution in a dynamic environment.


\section{Methods}

To address the aforementioned problems, we proposed a framework that consists of \textbf{Planning}, \textbf{Perception}, and \textbf{Execution}, as illustrated in Fig.~\ref{fig:overview}. 
This tight coupling between high-level planning, real-time scene interpretation, and low-level control constructs a closed-loop system that enables scalable and interpretable solutions for long-horizon assembly tasks.

\subsection{Preliminaries}
\textbf{\textit{System Setup:}}
The system setup consists of providing resources, specifying the domain, and initializing the world state, which are defined as follows:
\begin{itemize}
  \item \textbf{Resources Provision} $(V,O)$: $V$ is the human demonstration video. $O$ denotes the objects available, including both manipulable components and usable tools.
  \item \textbf{Domain Specification} $\mathcal{D}=(\mathcal{P}, \mathcal{C}, \mathcal{R}, \mathcal{A}, \mathcal{S})$: $\mathcal{P}$ is the set of unary predicates describing the object properties. Both $\mathcal{C}$ and $\mathcal{R}$ are sets of binary predicates, characterizing the inter-object constraints and relations, respectively. $\mathcal{A}$ is the set of available action primitives. $\mathcal{S}$ denotes the set of skill symbols, each of which represents a capability to accomplish a specific task. All elements in the domain vocabulary can be instantiated by being assigned to concrete objects. A full set of concrete examples is provided in Table~\ref{tab:domain_specification}.
  \item \textbf{World State Initialization} $\omega$: The world state is modeled as $\omega = (P, R)$, where $P,R$ comprise object properties and inter-object relations, respectively. In essence, $\omega$ captures the facts that hold in the environment. $\omega$ is initialized by loading a predefined specification of object properties and relations that characterize the task setup.
\end{itemize}

\textbf{\textit{Behavior Trees:}}
A BT is a directed rooted tree whose internal nodes serve as \textit{control flow nodes} and leaf nodes serve as \textit{execution nodes}, providing a hierarchical structure for task specification \cite{colledanchise2018behavior}. There are four types of nodes within this framework:
\begin{itemize}
    \item Sequence: A control flow node graphically represented as ``$\rightarrow$" that returns \texttt{Success} only if all children succeed; otherwise, it returns the status of the first child that returns \texttt{Failure} or \texttt{Running}.
    \item Selector (Fallback): A control flow node graphically represented as ``$?$" that returns the status of the first child that is \texttt{Success} or \texttt{Running}. It returns \texttt{Failure} only if all children fail.
    \item Condition: An execution node that evaluates a proposition, i.e., a constraint, a property, or a relation. It deterministically returns either \texttt{Success} or \texttt{Failure}.
    \item Action: An execution node that invokes an action defined as \(a_i(O_i)\), where \(a_i \in \mathcal{A}\) and \(O_i \subseteq O\). It returns \texttt{Running} while being executed asynchronously, \texttt{Failure} on error, and \texttt{Success} on completion. Upon success, its effects are updated to the world state.
\end{itemize}

A BT executes by recursively propagating ticks in a depth-first, left-to-right order. Tick propagation is typically repeated at a fixed control frequency.

When a condition node is placed under a selector, its success causes the selector to skip the remaining children, implying the target of the subsequent execution. When a condition node or a subtree is placed under a sequence, its success is required for the execution to proceed, making it function as a precondition. This framework follows the principle that every action node should be embedded in an action unit (e.g., the BT shown in Fig.~\ref{fig:lfd_prompt_pipeline}) that contains a target and optional precondition(s). 


\subsection{Planning: BT generation from video demonstration}

\begin{figure}
    \centerline{\includegraphics[width=1.0\linewidth]{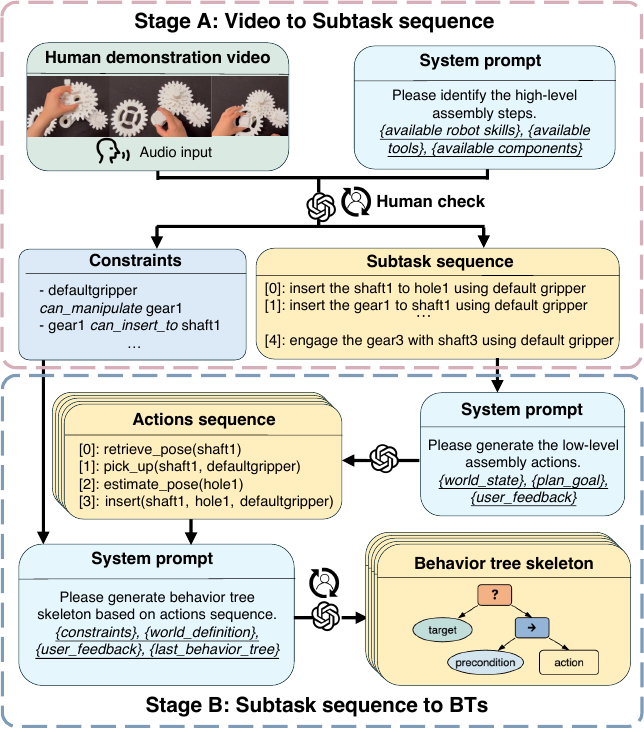}}
    \caption{\textbf{The planning module} involves two VLM-driven stages. Stage A generates subtasks and inter-object constraints by interpreting a human demonstration video; stage B structures and compiles the subtasks to generate the final BT. The human verification is also embedded in both generation stages to ensure the safety and reliability.}
    \label{fig:lfd_prompt_pipeline}
\end{figure}

The planning module leverages the generalization and reasoning capabilities of pre-trained VLMs to translate human-oriented instructional videos into robot-oriented BTs. Unlike the traditional role of an expert programmer, our module allows the operator to serve as a robot mentor, who simply demonstrates tasks in a human-centered manner, reviews the learned results, and provides minimal HITL guidance when necessary. As illustrated in Fig.~\ref{fig:lfd_prompt_pipeline}, the planning module operates in two stages: video interpretation and BT generation.



\textbf{\textit{Video interpretation:}}
Human demonstration videos are recorded from an action-centric view to capture a complete assembly procedure. Each video is accompanied by an auxiliary audio description, which is intentionally multilingual and stylistically diverse. Illustrative video samples are provided in Fig.~\ref{fig:video_sample}. To process the videos, the key frames and the audio are first extracted from the video. The key frames are selected manually to ensure that they capture the essential steps of the assembly. The extracted audio is then converted to text, providing a complementary prompt for further reasoning. Based on key frames and audio text, a VLM agent can infer the sequence of subtasks \(\langle \tau_1,\dots,\tau_N\rangle\) and constraints $C$ among relevant objects, grounded in the set of objects $O$ and available skills $\mathcal{S}$.
\begin{align}
    \tau_i &= (s_i, o_{i1}, o_{i2}), \; s_i \in \mathcal{S}, \\
    C &= \{c(o_1, o_2) \mid c \in \mathcal{C},\;o_1, o_2 \in O \},
\end{align}
where $\tau_i$ defines one subtask in the integrated assembly procedure and $C$ encodes intrinsic constraints among objects (e.g., $o_1$ \textit{can\_insert\_to} $o_2$), forming a crucial prerequisite for the BT generation.

\textbf{\textit{BT generation:}} The planning module generates BTs through a sequential process, where each subtree $b_i$ is created and verified individually. For each given subtask $\tau_i$, the process begins with an LLM agent planning an action sequence $\textbf{a}_i =\langle a_{i1}(O_{i1}),\dots, a_{iM}(O_{iM})\rangle$. Subsequently, as $\textbf{a}_i$ serves as a blueprint, $b_i$ is generated by an LLM agent, with reference to the domain $\mathcal{D}$, the constraints $C$, and the world state $\omega_i$ that describes the environment at the beginning of each subtask. The generated BT is then verified in a HITL process. Upon the generation of a correct subtree $b_i$, a BT simulator is utilized to perform a virtual tick, thereby transitioning the world state from $\omega_i$ to $\omega_{i+1}$. This resultant state $\omega_{i+1}$ serves as a prerequisite context for the synthesis of the subsequent subtree $b_{i+1}$. This iterative procedure is applied across the entire sequence of subtasks, culminating in the final BT sequence $\Pi=\langle b_0,\,\dots,\,b_N \rangle$.

\begin{figure}
    \centering
    \includegraphics[width=\linewidth]{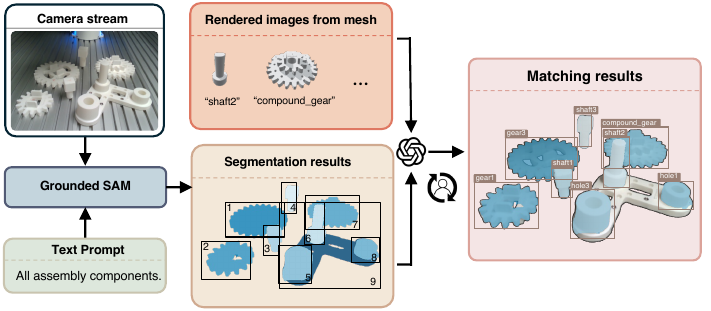} 
    \caption{\textbf{Objects recognition} includes two steps. First, the Grounded SAM~\cite{ren2024grounded} model is employed for coarse segmentation and numerical annotation of assembly components. Subsequently, a VLM matches these annotated masks with mesh-rendered images to obtain the final matched masks and their corresponding names.}
    \label{fig:gpt_matching_pipeline}
\end{figure}

\subsection{Perception: Semantic and dynamic scene understanding}
The perception module captures environmental changes caused by external disturbances through object recognition and continuous scene interpretation, allowing the world state $\omega^t$ to be updated accordingly and to remain a coherent representation of the dynamic environment.

\textbf{\textit{Object recognition}}
can be accomplished either through a VLM-based matching pipeline or by manual annotation. As illustrated in Fig.~\ref{fig:gpt_matching_pipeline}, the VLM-based pipeline leverages Grounded SAM \cite{ren2024grounded} as an object detector and the generalization capabilities of VLMs. Guided by a simple text prompt, the detector extracts object masks and annotates them with a number. These annotated masks, together with images rendered from the component meshes, are then provided to a VLM agent to match the scene objects with those in object set $O$, followed by a human check to ensure correctness. Alternatively, a web interface for manual annotation is developed that allows users to interactively mark each object in the scene corresponding to $O$.

\textbf{\textit{Runtime scene interpretation}} is achieved through two visual modules: (i) The first module $\mathcal{M}_1$ utilizes SAM2~\cite{ravi2024sam} to segment the scene based on the result of object recognition and maintain object tracking throughout the assembly process. In doing so, it produces masks of the objects, which provide an approximate representation of the object positions in the scene. (ii) The second module $\mathcal{M}_2$ utilizes FoundationPose~\cite{wen2024foundationpose} to maintain 6D-pose estimations of the objects involved in the current action $a_i(O_i)$, providing pose information for the executor.

\textbf{\textit{World state update}} is performed by an \textsc{UpdateState} operation, which applies the following rules based on $\mathcal{M}_1$ and $\mathcal{M}_2$ to update $\omega^t$ in runtime.
\begin{itemize}

    \item \textbf{Object position invariance.} For objects not involved in the current action \(a_i( O_i)\), any discrepancy in position identified by $\mathcal{M}_1$ will trigger an update of the recorded mask and position.
    \item \textbf{Relation validity.} For each relation \(r(o_i, o_j) \in R\), if a relative displacement of $o_i$ and $o_j$ is detected by $\mathcal{M}_1$, the relation is deemed invalid and removed from $R$, i.e., \(R^t \leftarrow R^{t-1} \setminus \{r(o_i, o_j)\}\).
    \item \textbf{Pose consistency.} For objects involved in the current action \(a_i(O_i)\), if $\mathcal{M}_2$ detects that an object \(o_j \in O_i\) is moved unexpectedly, the pose of $o_j$ should be considered unknown for the executor before the newly detected pose is retrieved. Accordingly, the object property is updated as \(P^t \leftarrow P^{t-1} \setminus \{\textit{pose\_is\_known}(o_j)\}\).
\end{itemize}
By applying these rules, the world state is continuously maintained as a coherent and reliable representation of the environment.

\subsection{Execution: Reactive BT with recovery mechanism}


The execution module carries out the planned BTs based on the real-time scene interpretation, enabling the assembly task to be accomplished in a dynamic environment. As illustrated in Algorithm~\ref{alg: algorithm}, each $b_i$ in the BT sequence $\Pi$ is sequentially extended and executed, with continuous world state monitoring and the recovery mechanism to ensure robustness.



\begin{figure}
\centering
\includegraphics[width=\linewidth]{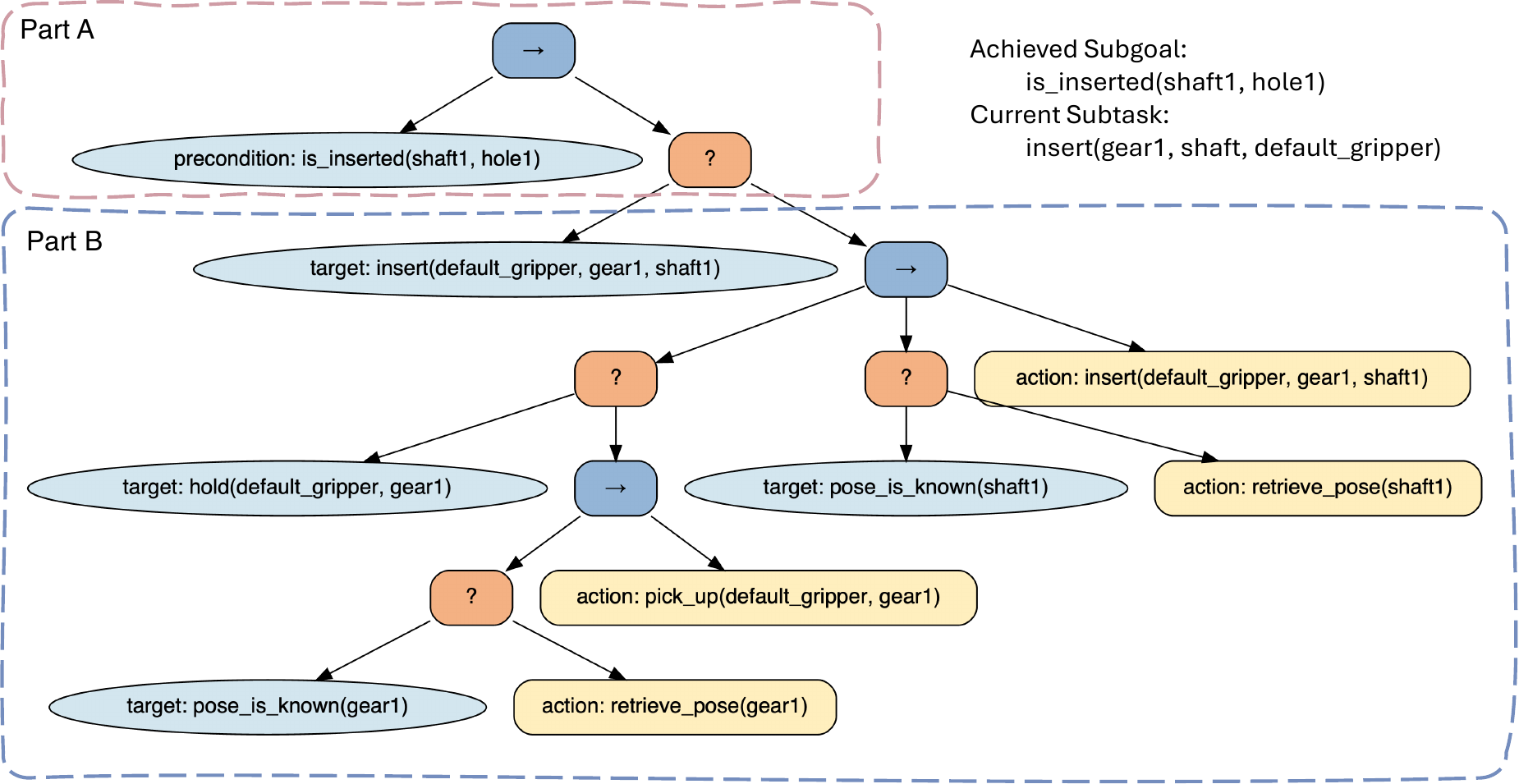} 
\caption{\textbf{Example of an extended executable BT} that highlights its two constituent parts, i.e., Part A: the extension mechanism; Part B: BT generated by a VLM for a subtask.}
\label{fig: bt_example}
\end{figure}

\textbf{\textit{BT extension:}} Each subtree $b_i$ corresponds to the subtask $\tau_i$, whose successful accomplishment yields an inter-object relation $r_i$ that represents the achieved subgoal (for instance, the accomplishment of \textit{insert(gripper, shaft, gear)} establishes the relation \textit{is\_inserted\_to(gear, shaft)}). During the execution of $b_i$, the validity of the relations established by previously executed subtrees must be preserved. To ensure this, these relations $(r_0,\dots, r_{i-1})$ are incorporated as preconditions of $b_i$ by placing them as the leading children of a sequence node, resulting in an extended BT $\hat{b_i}$ defined as \textsc{Sequence}$(r_0,\dots,r_{i-1}, b_i)$. An example of the extended BT is illustrated in Fig.~\ref{fig: bt_example}. 


\textbf{\textit{World state monitoring:}} Throughout the process, the world state $\omega^t$ is continuously updated by the operation \textsc{UpdateState}, while the BTs are executed sequentially in synchrony. The extended BT $\hat{b_i}$ is ticked at a frequency $f$ until success. At every tick, the condition nodes evaluate propositions against $\omega^t$, thereby verifying that both the previously achieved subgoals and the preconditions of the current action remain preserved.

\textbf{\textit{Recovery mechanism:}} When a precondition fails, the ongoing action will be suspended and the recovery mechanism will be triggered. The tick propagation will search for an action unit whose target effect can restore the violated condition. If such a unit exists, its action (e.g., retrieving pose) will be enforced before resuming the prior execution. Otherwise, self-recovery is impossible and the BT will return \texttt{Failure}; the mechanism thereby will identify the failed node and triggers replanning based on $\omega^t$: if a previously achieved subgoal $r_j$ with $j<i$ has been violated, the process will roll back to the stage $j$ and replan $b_j$; otherwise, replan the current subtree $b_i$.


Integrated with guarded preconditions, real-time evaluation, and corrective recovery, the system remains reactive to the dynamic environment, ensuring a robust execution for long-horizon tasks under disturbances.
\vspace{-0.5em}
\begin{algorithm}[h]
\footnotesize
\setlength{\baselineskip}{0.7\baselineskip}
\caption{Reactive Execution of BTs}
\SetKwComment{Comment}{/{\kern-0.2em}/ }{}
\newcommand{\SmallComment}[1]{\Comment*[r]{\scriptsize\sffamily{\kern-0.05em #1}}}
\KwIn{Subtree sequence $\Pi=\langle b_0,\dots,b_N\rangle$}
\KwData{Realtime World State $\omega^t$}
\label{alg: algorithm}
\vspace{0.8ex}
\textbf{Thread A (Reactive BT Execution Loop):}

\For{$i \gets 0$ \KwTo $N$}{
  \Repeat{\texttt{status} = \texttt{Success}}{
    
    $\hat{b_i} \gets$ \textsc{Sequence}$(r_0,\dots,r_{i-1}, b_i)$\;
    
    \texttt{status} $\gets$ \texttt{Running}\;
    
    \While{\texttt{status} = \texttt{Running}}{
      \texttt{status} $\gets$ \textsc{Tick}$(\hat{b_i},\omega^t,f)$\; \SmallComment{self-recovery embedded}
    }
    
    \If{\texttt{status} = \texttt{Failure}}{
      $n_{\mathrm{fail}} \gets$ \textsc{FailedNode}$(\hat{b_i})$\;
      
      \uIf{$n_{\mathrm{fail}} = r_j$ with $j < i$}{
        $b_j \gets$ \textsc{Replan}$(b_j,\omega^t)$\;
        
        $i \gets j$\;
        
        \textbf{break}\; \SmallComment{rollback to stage $j$}
      }
      \uElseIf{$n_{\mathrm{fail}} = b_i$}{
        $b_i \gets$ \textsc{Replan}$(b_i,\omega^t)$\; \SmallComment{replan the current BT}
      }
    }
  }
}

\BlankLine

\textbf{Thread B (world-state maintenance):}

\Repeat{\textsc{Thread A Ends}}{
  $a_i(O_i) \gets \textsc{CurrentAction()}$\;
  
  $\omega^t \gets \textsc{UpdateState}(\omega^{t-1}, a_i(O_i),\mathcal{M}_1, \mathcal{M}_2)$\;
  
}

\end{algorithm}


    
    
    
    
      
        
        



  
  


\section{Experiments and Results}
The proposed framework is specifically designed to address the challenges of complex, long-horizon tasks that involve multiple constraints and dynamic environmental uncertainties. Accordingly, the subsequent experiments are designed to address the following research questions:

\textbf{RQ1:} To what extent is our proposed video-to-BT planning module generalizable and reliable?

\textbf{RQ2:} How effectively can our perception module perform automated object recognition?

\textbf{RQ3:} How robust is our reactive BT in execution under external disturbances?

\subsection{Experiment Setup}

\begin{figure}[htb]
\centering
\vspace{-5pt}
\includegraphics[width=0.9\linewidth]{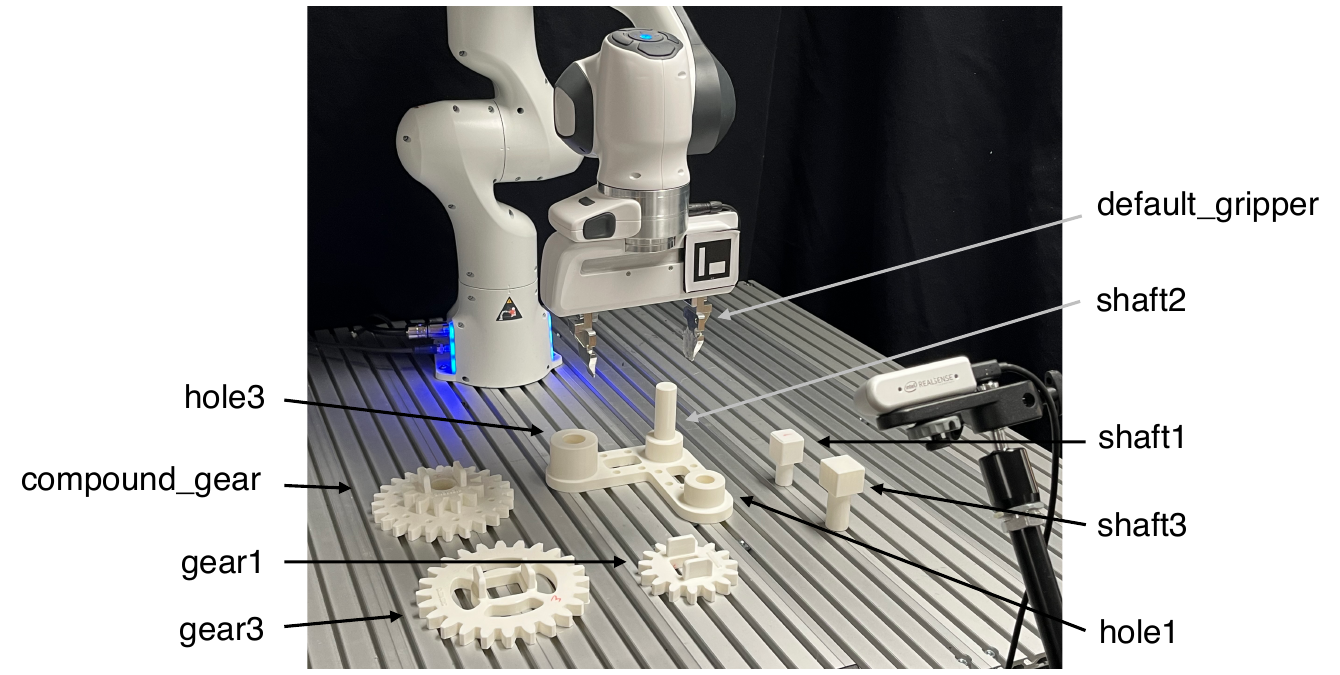}
\caption{\textbf{Experiment setup.} 
}
\label{fig:usecase_details}
\end{figure}

As illustrated in Fig.~\ref{fig:usecase_details}, the experiment setup consists of a Franka Emika Panda robot, an Intel Realsense D435i camera, and a gearset from the Siemens Robot Assembly Challenge. The fundamental action set employed by the BTs is derived from the taxonomy of skills described in~\cite{johannsmeier2025process}. The domain is specified in the Table~\ref{tab:domain_specification}.

\begin{table}[H]
\scriptsize 
\vspace{3pt}
\caption{Domain specification}
\vspace{-3pt}
\label{tab:domain_specification}
\setlength\tabcolsep{4pt} 

\begin{tabularx}{\columnwidth}{@{} c X @{}}
\toprule
\textbf{Category} & \textbf{Signature}\\
\midrule
$\mathcal{P}$ & is\_empty, pose\_is\_known \\
$\mathcal{C}$ & can\_manipulate, can\_insert\_to, can\_engage\_with, can\_place\_on \\
$\mathcal{R}$ & is\_inserted\_to, is\_engaged\_with, is\_placed\_on, hold \\
$\mathcal{A}$ & change\_tool, pick\_up, put\_down, insert, engage, place, retrieve\_pose \\
$\mathcal{S}$ & insert, engage, place \\
\bottomrule
\end{tabularx}

\vspace{-2pt}
\end{table}


\begin{figure*}[t]
  \centering
  \includegraphics[width=1.0\textwidth]{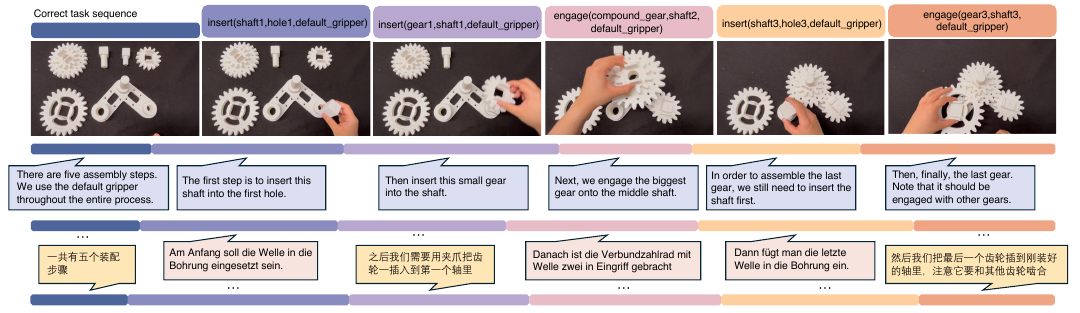}
  \caption{\textbf{Human demonstration video samples.} The videos showcase assembly procedures, supplemented by flexible audio descriptions in multiple languages (e.g., English, German, and Chinese) and varied narrative styles.}
  \label{fig:video_sample}
\end{figure*}

\subsection{BT generation from demonstration (\textbf{RQ1})}

\textit{\textbf{Experiment Design:}}
To facilitate a comprehensive evaluation of our planning module, 25 human demonstration videos have been prepared. The accompanying instructions are intentionally diverse, spanning multiple languages and varied narrative styles (samples in Fig.~\ref{fig:video_sample}). The evaluation assesses the module from three key aspects: its generalization to diverse video inputs, the reliability endowed by the HITL process, and its compatibility with varying foundation models. For each model tested, our analysis distinguishes between the performance at two critical stages: the initial response from direct prompting and the refined response following a single round of human feedback.

The quality of generated BTs from a single video is then quantified using the following three metrics, designed to assess both the semantic correctness of its decomposed subtasks as well as the logical and syntactic integrity of the generated BTs:

\begin{itemize}
    \item Task Decomposition Accuracy (TDA). The percentage of generated subtasks that correctly correspond to the video. A subtask is deemed correct if its semantic elements (skill, components, and tool) are consistent with the video demonstration.
    
    \item Logical Coherence Rate (LCR): The percentage of generated BTs that exhibit a logically coherent sequence of actions. A BT is considered coherent if it adheres to procedural constraints and preconditions, such as requiring an object to be grasped before it can be placed.

    \item Syntactic Validity Rate (SVR): The percentage of generated BTs that are syntactically valid. This metric verifies that the BT structure adheres to its formal grammar and formatting rules.
    
\end{itemize}

\textit{\textbf{Results:}}
The evaluation results are summarized in the Table \ref{tab:lfd_accuracy}, presenting the average rate for each metric calculated across all 25 test videos.

\begin{table}[H]
\scriptsize 
\vspace{4pt}
\caption{BT generation result using different VLMs}
\vspace{-3pt}
\label{tab:lfd_accuracy}
\setlength\tabcolsep{1.5pt} 
\begin{tabular*}{\columnwidth}{@{\extracolsep{\fill}} l ccc ccc ccc}
\toprule
 & \multicolumn{3}{c}{\textbf{GPT-4o}} & \multicolumn{3}{c}{\textbf{Gemini-1.5-flash}} & \multicolumn{3}{c}{\textbf{Qwen-2.5VL-72b}} \\
\cmidrule(r){2-4} \cmidrule(lr){5-7} \cmidrule(l){8-10}
\textbf{Method} & \textbf{TDA} & \textbf{LCR} & \textbf{SVR} & \textbf{TDA} & \textbf{LCR} & \textbf{SVR} & \textbf{TDA} & \textbf{LCR} & \textbf{SVR} \\
\midrule
Initial response & 98\% & 99\% & 100\% & 81\% & 61\% & 99\% & 76\% & 80\% & 100\% \\
Refined response & 100\% & 100\% & 100\% & 94\% & 88\% & 100\% & 79\% & 100\% & 100\% \\
\bottomrule
\end{tabular*}

\vspace{-8pt}
\end{table}
The results confirm the ability of the VLM-based planning module to generate high-quality BTs by combining high-performance autonomous generation with a robust HITL safeguard. In task decomposition, initial TDA scores were impacted by visual ambiguities such as poor component visibility or ambiguous verbal instructions. However, a single refinement proved highly effective, raising the accuracy for GPT-4o to 100\% and Gemini-1.5-flash to 94\%. This process successfully corrects initial imperfections to ensure an accurate final task sequence.

For the subsequent BT structural generation provided with the correct task sequence, GPT-4o achieved an excellent initial LCR of 99\%. The HITL process also bridged performance gaps for other smaller models, elevating the LCRs of Gemini-1.5-flash from 61\% to 88\% and Qwen-2.5VL-72b from 80\% to 100\%. This demonstrates the module's capacity to consistently deliver logically coherent and executable BTs.

\subsection{Semantic perception (\textbf{RQ2})}

The perception module allows both manual annotation and a VLM-based matching pipeline as illustrated in Fig.~\ref{fig:gpt_matching_pipeline} for object recognition. Manual annotation consistently provides correct and stable recognition. As object recognition constitutes the foundation of subsequent scene perception, the experiment aims to examine how reliably the VLM-based matching pipeline can substitute for the manual annotation, with GPT-4o employed as the VLM agent.

\textit{\textbf{Experiment Design:}} This experiment evaluates how two key factors influence the success rate of the mask-to-name matching process. The first factor is the number of interaction rounds that we assign to 0, 1, 3, and 5 to represent varying levels of human assistance. The second is the composition of the human feedback, where we vary the number of correct mask-name pairs (e.g., one versus multiple) provided in each round to represent different information densities. For each experimental configuration, we measure the overall success rate, defined as the proportion of 15 complete trials that result in a perfectly correct match for all assembly components. The aggregated results are presented in Table \ref{tab:gpt_matching}.

\begin{table}[H]
\scriptsize 
\vspace{2pt}
\caption{Accuracy rate of GPT-4o matching}
\vspace{-3pt}
\label{tab:gpt_matching}
\setlength\tabcolsep{1.5pt} 

\begin{tabular*}{\columnwidth}{@{\extracolsep{\fill}} l cccc}
\toprule
& \multicolumn{4}{c}{\textbf{Number of Refinement Steps}} \\
\cmidrule(lr){2-5}
\textbf{Information Density} & \textbf{0} & \textbf{1} & \textbf{3} & \textbf{5} \\
\midrule
one pair & 4/15 & 6/15 & 12/15 & 15/15 \\
multiple pairs & 4/15 & 13/15 & 15/15 & 15/15 \\
\bottomrule
\end{tabular*}

\vspace{-8pt}
\end{table}

\textit{\textbf{Results:}} Table \ref{tab:gpt_matching} indicates that iterative design of the pipeline is robust, consistently reaching a 15/15 correct match within five cycles. It is also observed that more informative feedback accelerates this process, with richer prompt details leading to faster convergence. Overall, these findings confirm that the automated pipeline performs on par with manual annotation and serves as a stable alternative.

\begin{figure*}[t]
  \centering
  \includegraphics[width=\linewidth]{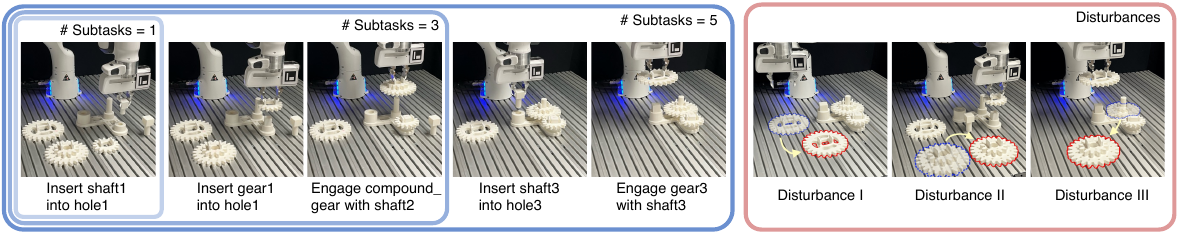}
  \captionof{figure}{\textbf{Demonstration of the assembly procedure and disturbances}. \textbf{Left}: Assembly execution procedures consisting of one, three, and five subtasks. With the growth of task length, the number of involved objects also increases, making the overall execution more challenging. \textbf{Right}: Examples of external disturbances: \textit{Disturbance I}, moving the gear3 while the robot is approaching it; \textit{Disturbance II}, moving the compound\_gear when the robot tries to insert the gear1 into the shaft1; \textit{Disturbance III}, extracting the compound\_gear from the shaft2 before the robot attempts to insert the gear3 into the shaft3.}
  \label{fig:bt_exe_snapshots}
\end{figure*}

\begin{figure*}[t]
  \centering
  \includegraphics[width=\linewidth]{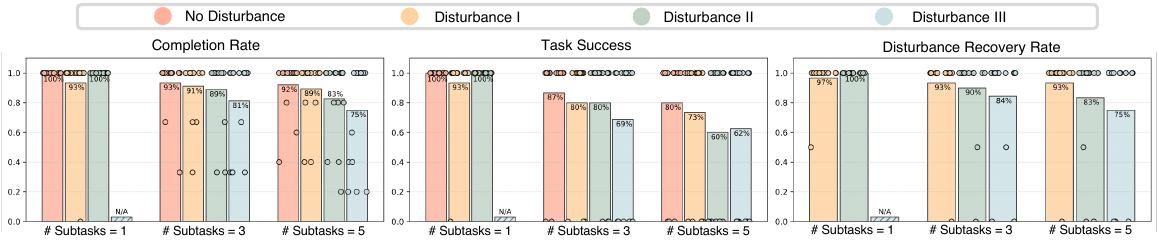}
  \captionof{figure}{\textbf{Execution results}. Performance comparison across varied task lengths and disturbance types. The bar heights indicate the average value, with individual trial results shown as scatter points.}
  \label{fig:bt_exe_results}
\end{figure*}

\subsection{Real-world BT Execution (\textbf{RQ3})}
Provided with the correctly planned BTs and segmentation results from the previous stages, the execution of the BTs is then deployed on the Franka robot to validate the robustness and reactivity of our framework in executing assembly tasks in the dynamic environment.

\textit{\textbf{Experiment Design:}}
As illustrated in Fig.~\ref{fig:bt_exe_snapshots}, the experiments are designed with three difficulty levels, characterized by assembly tasks consisting of one, three, and five subtasks. In addition, four cases are considered to verify the disturbance resilience of the system: \textit{No external disturbance}; \textit{Disturbance I}, displacement of the component involved in the current action; \textit{Disturbance II}, displacement of components that have not yet been involved in any actions; and \textit{Disturbance III}, violation of a previously achieved goal. Each experimental setting was run for 15 trials.

The following quantitative evaluation metrics are measured for each single trial:
\begin{itemize}
    \item Task Success (TS): Whether the entire assembly task is completed successfully, requiring full execution of all actions and proper handling of perturbations.
    \item Completion Rate (CR): The percentage of accomplished subtasks.
    \item Disturbance Recovery Rate (DRR): The percentage of disturbances that are successfully addressed.
\end{itemize}

\textit{\textbf{Results:}} The results are shown in Fig.~\ref{fig:bt_exe_results}. Without disturbance, both average CR and average TS among the 15 trials remain high, above 90\% and 80\% respectively, confirming that the system can reliably execute long-horizon tasks. Under disturbances, the performance degrades moderately: the average CR exceeds 75\%  and the average TS exceeds 60\% for all cases, highlighting the robust performance and effective recovery under diverse disturbances.

To further understand the disturbance robustness of the system, an analysis of failed recovery cases was conducted:
 
\begin{itemize}
    \item \textit{Disturbance I} is handled based on precise pose tracking of a single object at the primitive action level. The main reason for the failed handling stems from occasional estimation deviations, which cause execution errors in individual actions.
    \item \textit{Disturbance II} and \textit{Disturbance III} are handled based on the object tracking throughout the entire process. Nevertheless, the average DRR exhibits a noticeable decline from 100\% to 83\% and from 84\% to 75\%, respectively, as the number of subtasks and involved objects increases, reflecting the rising likelihood of tracking losses or missassignments. Such errors cause misalignments between the scene graph and the physical environment, leading to incorrect condition evaluations and ultimately execution failure.
\end{itemize}

The failure case analysis points to the limitation of our framework: perception-related errors occur more frequently as the task length increases, which causes a decrease in success rate for longer horizons; but it also suggests that advances in perception modules will enhance the robustness and scalability of the entire system.

\section{Conclusions}

This paper presents \textit{Video-to-BT}, a holistic framework enabling non-experts to deploy robots for complex, long-horizon assembly tasks. Our approach effectively bridges the gap between high-level planning and real-world execution by translating demonstration videos into executable BTs using foundation models. The core of our framework lies in the tight coupling of three key modules. First, a robust planning phase generates executable BTs. The execution of these BTs is then continuously guided by a semantic perception module that detects real-time disturbances. This synergy between planning and perception empowers the robot to execute tasks reactively and robustly in dynamic environments. The experiments confirm the effectiveness of the framework in reliable planning and robust execution. 
In future work, we will focus on extending the framework’s applicability to a broader range of manipulation tasks.


\newpage

\bibliography{IEEEabrv,mybib2022}
\bibliographystyle{myIEEEtran}


\label{last-page}
\end{document}